\documentclass[10pt,journal,compsoc,twocolumn]{IEEEtran}
\usepackage{amsmath,epsfig,multirow,multicol,float}
\usepackage{subfigure}
\usepackage{graphicx}
\usepackage{psfrag,cite}
\usepackage{amsmath,amssymb,epsfig,dsfont}
\usepackage{color}
\begin{document}
\graphicspath{{figures/}}
\newcommand{\eqs}{\small}
\newcommand{\sps}{\scriptsize}%
\title{On the Convergence of
Extended Variational Inference for Non-Gaussian Statistical Models}

\author{Zhanyu~Ma, Jalil~Taghia, and~Jun~Guo
}

\IEEEcompsoctitleabstractindextext{
\begin{abstract}
Variational inference (VI) is a widely used framework in Bayesian estimation. For most of the non-Gaussian statistical models, it is infeasible to find an analytically tractable solution to estimate the posterior distributions of the parameters. Recently, an improved framework, namely the extended variational inference (EVI), has been introduced and applied to derive analytically tractable solution by employing lower-bound approximation to the variational objective function. Two conditions required for EVI implementation, namely the weak condition and the strong condition, are discussed and compared in this paper. In practical implementation, the convergence of the EVI depends on the selection of the lower-bound approximation, no matter with the weak condition or the strong condition. In general, two approximation strategies, the single lower-bound (SLB) approximation and the multiple lower-bounds (MLB) approximation, can be applied to carry out the lower-bound approximation. To clarify the differences between the SLB and the MLB, we will also discuss the convergence properties of the aforementioned two approximations. Extensive comparisons are made based on some existing EVI-based non-Gaussian statistical models. Theoretical analysis are conducted to demonstrate the differences between the weak and the strong conditions. Qualitative and quantitative experimental results are presented to show the advantages of the SLB approximation.
\end{abstract}

}
\maketitle

\IEEEdisplaynotcompsoctitleabstractindextext

\IEEEpeerreviewmaketitle

\section{Introduction}
\vspace{0mm}
\label{sec:intro}
\IEEEPARstart{G}aussian distribution is the ubiquitous probability distribution used in statistics, signal processing, and pattern recognition areas~\cite{Park2013}. However, not all the data being processed are Gaussian distributed~\cite{Ma2011}. In many real-life applications, the distribution of data is asymmetric and, therefore, is not Gaussian distributed~\cite{Nguyen2013}. For example, the image pixel values~\cite{Ma2011a,Bouguila2006}, the reviewer's rating to an item in a recommendation system~\cite{Ma2015,Salakhutdinov2008,Salakhutdinov2008a}, and the DNA methylation level data~\cite{Ji2005} are distributed in a range with bounded support. The diversity gain over the $K_G$ fading~\cite{Jung2014} and the periodogram coefficients in speech enhancement~\cite{Mohammadiha2013,Mohammadiha2013a} are semi-bounded (nonnegative). The spatial fading correlation~\cite{Mammasis2009} and the yeast gene expressions~\cite{Taghia2014} have directional property so that the $l_1$ norm equals one. In signal processing, the acoustic noise with colored spectra~\cite{Zao2012} and the measurement noise in state-space model~\cite{Xu2014} are heavy-tailed. In the stock market, the asymptotic behavior of the first-order autoregressive (AR) process is clearly non-Gaussian~\cite{Amini2013} and the underlying Bayesian copula model for the stock index series are non-Gaussian as well~\cite{Xu2015}. Although the above mentioned data represent diverse characteristics, a common property is that, these data~\emph{not only} have specific support range,~\emph{but also} have non-bell distribution shape. The natural properties of Gaussian distribution (the definition domain is unbounded and the distribution shape is symmetric) do not fit such data well. Hence, these data are non-Gaussian distributed. It has been found in recent studies that explicitly utilizing the non-Gaussian characteristics can significantly improve the practical performance~\cite{Ma2011,Ma2011a,Bouguila2006,Ji2005,Jung2014,Mohammadiha2013,Mammasis2009,Mohammadiha2013a,Taghia2014,Zao2012,Xu2014}. Hence, it is of particular importance and interest to make thorough studies of the non-Gaussian data and non-Gaussian statistical models.

Bayesian analysis plays an essential role in parameter estimation of statistical models~\cite{Fukunaga1990,Jain2000,Bishop2006,Bernardo2000}. Unlike the conventionally used maximum-likelihood (ML) estimation~\cite{Dempster1977}, Bayesian estimation assumes that the parameters potentially follow underlying distributions and derives the posterior distributions of the parameters by applying the Bayes' theorem~\cite{Stigler1982} through combining the prior distributions with the likelihood function obtained from the observed data~\cite{Bishop2006,Tipping2004}. Estimation of the posterior distribution via Bayesian estimation has several advantages over the ML estimation. Firstly, it gives statistical description to the parameters, rather than the simple point estimate that yield by the ML estimation. This makes Bayesian estimation more robust and reliable, by including the resulting uncertainty into the estimation~\cite{Bernardo2000}. Secondly, it can potentially prevent the overfitting problem, which is one of the drawbacks the ML estimation suffers. This is mainly due to the advantage of Occam's razor effect in Bayesian estimation. Last but not the least, Bayesian estimation can estimate the model complexity automatically from the data. In ML estimation, model complexity decision usually requires cross validations and, therefore, is computationally costly~\cite{Bishop2006,Dempster1977}.

Varitional inference (VI) framework, among others, is a widely used strategy to infer the posterior distribution of the parameters in Bayesian analysis~\cite{Bishop2006,Jordan1999,Blei2005,Fox2012}. In a full Bayesian model where all the parameters are assigned with prior distributions, we minimize the Kullback-Leibler (KL) divergence of the true posterior distribution from the approximating one to obtain an optimal approximation to the posterior distribution~\cite[Ch. 10]{Bishop2006}. This procedure is equivalent to maximizing the lower-bound to the marginal likelihood (model evidence). The optimal posterior distribution can be obtained by iteratively updating one variable (or one variable group) while fixing the rest. However, unlike the famous Gaussian distribution~\cite{Bishop2006,Ormerod2012,Park2013}, most of the non-Gaussian statistical models (e.g., beta mixture model (BMM)~\cite{Bouguila2006, Ma2011a}, Dirichlet mixture model (DMM)~\cite{Fan2012}, von-Mises Fisher mixture model (VMM)~\cite{Taghia2014}, beta-Gamma nonnegative matrix factorization (BG-NMF)~\cite{Ma2015}) do not have analytically tractable solution to estimate the posterior distribution of the parameters. Numerical methods,~\emph{e.g.}, Newton-Raphson algorithm, Gibbs sampling, Markov Chain Monte Carlo, are usually employed to sample from the posterior distribution~\cite{Blei2005,Bouguila2006}. Numerical method often depends on Markov chain convergence and is in general computationally costly, especially in the high-dimensional space~\cite{Blei2004}.

Recently, an improved framework, namely the extended variational inference (EVI)~\cite{Blei2006,Hoffman2010,Braun2010,Ma2011a,Fan2012,Taghia2014,Ma2015}, has become popular in solving the above mentioned problem. Similar as the VI framework, EVI also seeks an optimal approximation to the posterior distribution. The difference is that EVI relaxes the objective function (the evidence lower-bound to the marginal likelihood) by constructing lower-bound approximation to the objective function. This lower-bound relaxation, which uses the convexity or relative convexity~\cite{Boyd2004} of the objective function, can yield analytically tractable solution so that the parameter estimation is facilitated. Although systematic bias has been introduced due to the lower-bound approximation, several works have demonstrated the advantages of EVI in Bayesian estimation of statistical models~\cite{Hoffman2010,Ma2011a,Taghia2014,Ma2015}. In Bayesian estimation of BMM, Ma et al.~\cite{Ma2011a} derived an analytically tractable solution which outperforms the numerical Gibbs sampling based method~\cite{Bouguila2006}. As an extension work, Bayesian estimation of DMM via EVI have been proposed in~\cite{Fan2012}, respectively. For directional data, von-Mises Fisher distribution is an important model in several applications. Analytically tractable solution to Bayesian estimation of VMM has been proposed by using EVI to provide lower-bound approximation~\cite{Taghia2014}. For non-negative matrix factorization (NMF), EVI was also applied in deriving analytically tractable solutions for Poisson process (discrete) NMF~\cite{Cemgil2009}, Gamma process NMF in music recording~\cite{Hoffman2010}, and beta-Gamma NMF for bounded support data~\cite{Ma2015}.

Convergence is an important issue in parameter estimation algorithm. For VI-based method, the objective function maximized during each iteration is convex or relatively convex in terms of the target variable's posterior distribution~\cite{Bishop2006,Boyd2004}. Hence, convergence is theoretically guaranteed. In EVI, the introduced lower-bound approximation to the objective function can be obtained either via a single extension over the whole variable group or multiple extensions, one for a subset of the whole variable group. Based on this, two lower-bound approximation strategies are obtained, one is the single lower-bound (SLB) approximation~\cite{Hoffman2010,Taghia2014,Ma2015} and the other is the multiple lower-bounds (MLB) approximation~\cite{Ma2011a,Fan2012}. For EVI with SLB approximation, convergence is also guaranteed because the original objective function is replaced by one single lower-bound and this new objective function (\emph{i.e.}, the single lower-bound to the original objective function in VI) is convex (or relatively convex) and maximized during each iteration. However, when applying EVI with MLB approximation, the variable group is divided into different disjoint subsets and there exists different lower-bound approximations to the objective function. During each iteration, different lower-bounds, one for each variable subset, are maximized iteratively. Since the new objective function is not unique, convergence can not be theoretically guaranteed.

In order to clarify the convergence property of the EVI framework, we will discuss and summarize the conditions that required in EVI implementation. The SLB and MLB approximations will also be analyzed and compared qualitatively and quantitatively. Experimental results based on the recently proposed EVI-based BMM and DMM estimation algorithms will be presented to demonstrate the advantages of the SLB approximations. We draw some conclusions in the end.

\section{Variational Inference and Extended Variational Inference}
\vspace{0mm}
\subsection{Variational Inference}
\vspace{0mm}
In Bayesian estimation, a universal solution to the variational inference (VI) framework~\cite{Jordan1999} is to approximate the posterior distribution by a product of several factor distributions and then update each factor distribution individually~\cite{Bishop2006}. This method is the so-called factorized approximation (FA) which was developed from the mean field theory in physics~\cite{Jaakkola2001}. With the FA method, the variational objective function that we want to maximize can be represented as the negative KL divergence as
\begin{equation}
\eqs
\label{Eq: Lowerbound}
\mathcal{L} = \mathbf{E}_{\mathbf{Z}}\left[{\ln p(\mathbf{X},\mathbf{Z})} - \ln  q(\mathbf{Z})\right],
\end{equation}
where $\mathbf{X}$ is the observed data, $\mathbf{Z}$ denotes all the random variables. If  $\mathbf{Z}$ can be (approximately) factorized into $M$ disjoint groups as $\mathbf{Z} = \left\{\mathbf{Z}_1,\ldots,\mathbf{Z}_i,\ldots,\mathbf{Z}_M\right\}$ and we approximate the true posterior distribution $p(\mathbf{Z}|\mathbf{X})$ as
\begin{equation}
\eqs
p(\mathbf{Z}|\mathbf{X}) \approx q(\mathbf{Z}) = \prod_{i=1}^M q_i(\mathbf{Z}_i),
\end{equation}
the optimal solution can be written as
\begin{equation}
\eqs
\label{Eq: Optimal Solution}
\ln q_i^*(\mathbf{Z}_i) = \mathbf{E}_{\mathbf{Z}\backslash_{\mathbf{Z}_i}}\left[\ln p(\mathbf{X},\mathbf{Z})\right] + \text{const}.
\end{equation}
The operator $\mathbf{E}_{\mathbf{Z}\backslash_{\mathbf{Z}_i}}$ means expectation with respect all the variables in $\mathbf{Z}$, except for $\mathbf{Z}_i$. If the optimal solution to the posterior distribution of $\mathbf{Z}_i$, which is $\ln q_i^*(\mathbf{Z}_i)$ in~\eqref{Eq: Optimal Solution}, has the same logarithmical form as the prior distribution, the conjugate match between the prior and the posterior distributions are satisfied. Then we have obtained an analytically tractable solution. However, this conjugate match is not satisfied in most of the practical problems~\cite{Ma2011a,Fan2012,Ma2015}. This is due to the fact that the optimal solution depends on the expectation computed with respect to the factor distribution~\cite{Bishop2006}.
\vspace{0mm}
\subsection{Extended Variational Inference}
\vspace{0mm}
\label{Sec: EVI}
In order to satisfy the conjugate match requirement, some approximations can be applied to get a nearly optimally analytically tractable solution. Braun et al.~\cite{Braun2010} considered the zeroth-order and first-order delta method for moments~\cite{Bickel2007} to derive an alternative for the objective function to simplify the calculation. Blei et al.~\cite{Blei2006} proposed a correlated topic model (CTM) and used a first-order Taylor expansion to preserve a bound such that an intractable expectation was avoided. Similar idea was also applied in~\cite{Ma2011a,Fan2012,Taghia2014} for approximating the posterior distributions in BMM, DMM, and VMM, respectively. Using Jensen's inequality has become commonplace in variational inference. In~\cite{Hoffman2010}, the concavity of the function $-x^{-1}$ and the convexity of $-\log x$ were studied and the Jensen's inequality and the first-order Taylor expansion were applied to approximately calculated the posterior distribution. Moreover, the EVI strategy was also applied in low rank matrix approximation area~\cite{Ma2015}, where the Taylor expansion and Jensen's inequality were both applied for the purpose of deriving analytically tractable solution.
\begin{table*}[!t]
  \vspace{0mm}
  \caption{\label{Tab: Required Conditions for EVI} \footnotesize Required conditions for EVI.}
 \centering
\sps
 \begin{tabular}{|c|c|c|c|}
 \hline
                  & Auxiliary function & Form of the Auxiliary Function & Systematic Gap\\
 \hline
 Strong condition & ${p}(\mathbf{X},\mathbf{Z})\geq \widetilde{p}_{s}(\mathbf{X},\mathbf{Z})$ & \multirow{2}{*}{$\mathbf{E}_{\mathbf{Z}\backslash_{\mathbf{Z}_i}}\left[\ln \widetilde{p}(\mathbf{X},\mathbf{Z})\right] \approxeq \ln {p}_i(\mathbf{Z}_i) $ $^{\dag}$} & \multirow{2}{*}{$\mathcal{G}_{\text{s}}>\mathcal{G}_{\text{w}}$}\\
 \cline{1-2}
 Weak condition   & $\mathbf{E}_{\mathbf{Z}}\left[\ln p(\mathbf{X},\mathbf{Z})\right] \geq \mathbf{E}_{\mathbf{Z}}\left[\ln \widetilde{p}_{w}(\mathbf{X},\mathbf{Z})\right]$ & & \\
\hline
\end{tabular}\\
$^{\dag}$ ``$\approxeq$'' denotes that the two formulations at the LHS and RHS have the same mathematical form, up to a constant difference.
\vspace{-5mm}
\end{table*}
All the aforementioned works utilized the following property.
Given an auxiliary function $\widetilde{p}(\mathbf{X},\mathbf{Z})$ which satisfies
\begin{equation}
\eqs
\label{Eq: Auxiliary Function}
\mathbf{E}_{\mathbf{Z}}\left[\ln p(\mathbf{X},\mathbf{Z})\right] \geq \mathbf{E}_{\mathbf{Z}}\left[\ln \widetilde{p}(\mathbf{X},\mathbf{Z})\right],
\end{equation}
the variational objective function (see~\cite{Bishop2006}, pp. 465 for more details) can be lower-bounded as
\begin{equation}
\eqs
\begin{split}
\label{Extended FA}
\mathcal{L} =&\mathbf{E}_{\mathbf{Z}}\left[\ln p(\mathbf{X},\mathbf{Z})\right] - \mathbf{E}_{\mathbf{Z}}\left[\ln q(\mathbf{Z})\right]\\
\geq &\mathbf{E}_{\mathbf{Z}}\left[\ln \widetilde{p}(\mathbf{X},\mathbf{Z})\right] - \mathbf{E}_{\mathbf{Z}}\left[\ln q(\mathbf{Z})\right]\\
\triangleq &\mathcal{\widetilde{L}}.
\end{split}
\end{equation}
Then we can maximize $\mathcal{\widetilde{L}}$, which is an lower-bound to the original objective function $\mathcal{\widetilde{L}}$, to asymptotically reach the maximum value of $\mathcal{{L}}$~\cite{Hoffman2010,Ma2011a,Fan2012}. The approximated optimal solution in this case is written as
\begin{equation}
\eqs
\label{Eq: Optimal Approximating Solution}
\ln \widetilde{q}_i^*(\mathbf{Z}_i) = \mathbf{E}_{\mathbf{Z}\backslash_{\mathbf{Z}_i}}\left[\ln \widetilde{p}(\mathbf{X},\mathbf{Z})\right] + \text{const}.
\end{equation}
This method is the so-called EVI framework~\cite{Blei2006,Hoffman2010,Braun2010,Ma2011a,Fan2012,Taghia2014,Ma2015}. Although it introduces systematic gap when involving the lower-bound approximation, the EVI allows more flexibility when calculating intractable integrations in non-Gaussian statistical models and provides a convenient way to obtain an analytically tractable solution.
\vspace{0mm}

\section{Convergence of EVI}
\vspace{0mm}
\subsection{Weak Condition and Strong Condition}
\vspace{0mm}
\label{Sec: Required Conditions}

\begin{figure}[!t]
\vspace{0mm}
\psfrag{A}[][]{\tiny $p(\mathbf{X},\mathbf{Z})$}
\psfrag{B}[][]{\tiny $\widetilde{p}_{\text{w}}(\mathbf{X},\mathbf{Z})$}
\psfrag{C}[][]{\tiny $\widetilde{p}_{\text{s}}(\mathbf{X},\mathbf{Z})$}
\psfrag{D}[][]{\tiny $\mathcal{G}_{\text{w}}$}
\psfrag{E}[][]{\tiny $S_1$}
\psfrag{F}[][]{\tiny $S_2$}
\psfrag{G}[][]{\tiny $S_3$}
     \centering
          \subfigure[\label{Subfig: Weak}\sps Weak condition of EVI.]{\includegraphics[width=.235\textwidth]{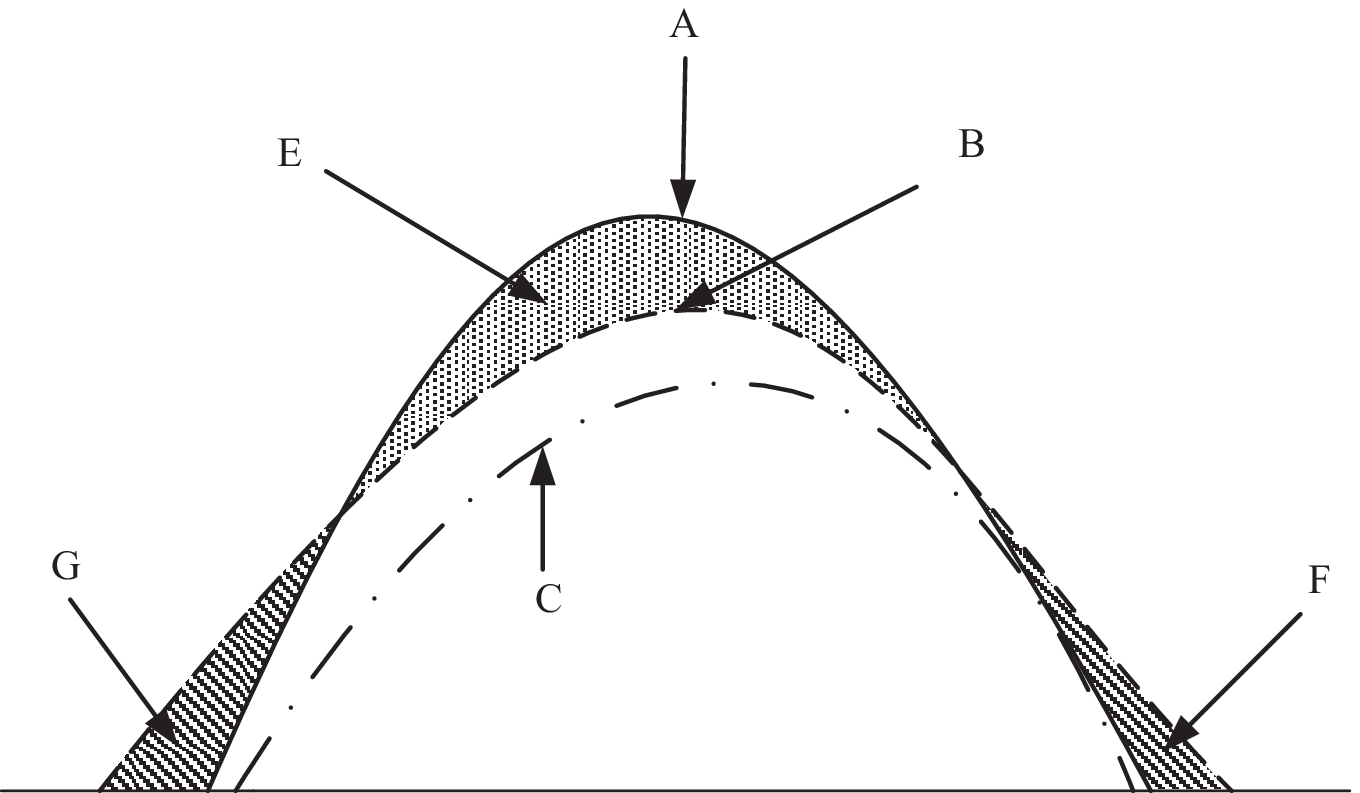}}\hspace{2mm}
          \subfigure[\label{Subfig: Strong}\sps Strong condition of EVI.]{\includegraphics[width=.235\textwidth]{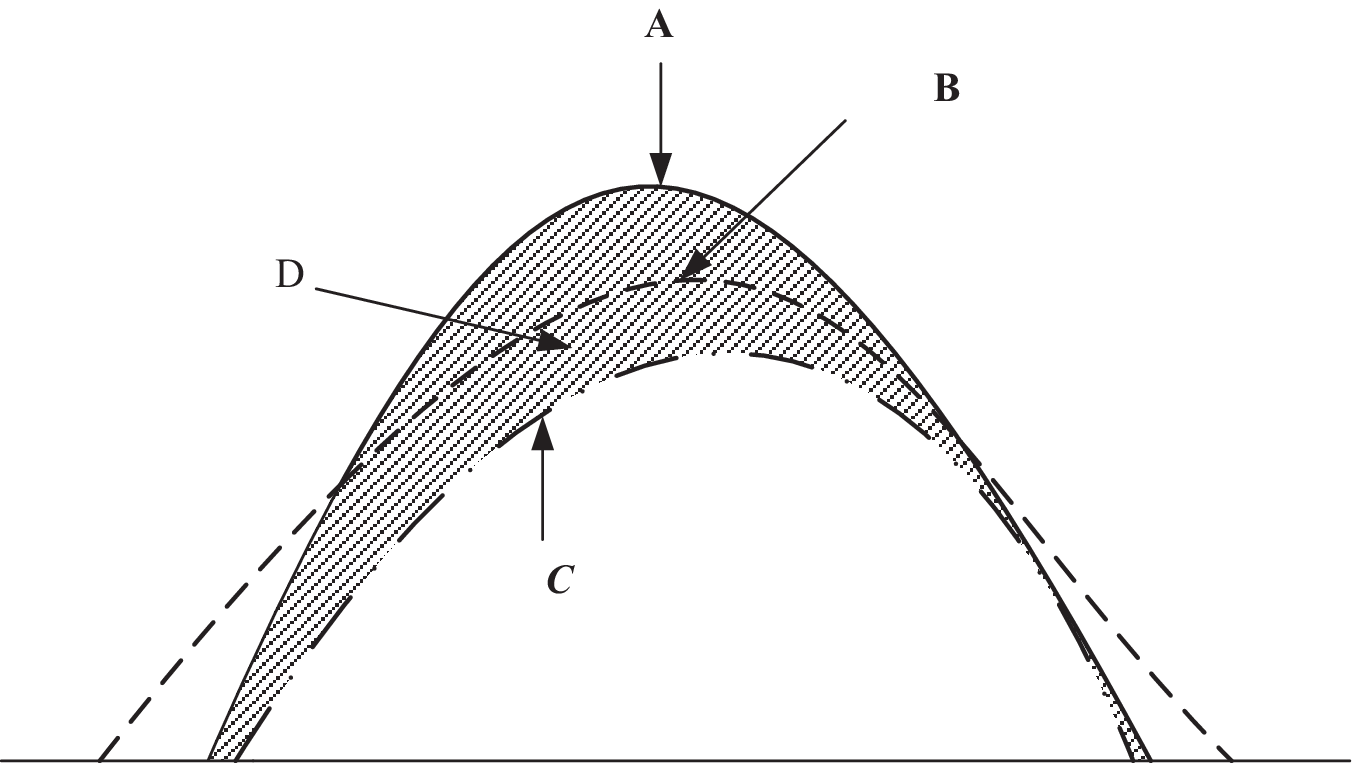}}\hspace{2mm}
\vspace{0mm}
          \caption{ \label{Fig: WeakAndStrongConditions}\footnotesize Comparisons of the weak and the strong conditions of EVI. The systematic gap introduced by the weak condition can be calculated as $\mathcal{G}_{\text{w}}=S_1-(S_2+S_3)$. For either the strong or the weak condition, the auxiliary function is chosen to minimize the gap as much as possible. Generally speaking, the systematic gap $\mathcal{G}_{\text{w}}$ is smaller than $\mathcal{G}_{\text{s}}$.}
\vspace{-5mm}
\end{figure}

As mentioned in Sec.~\ref{Sec: EVI}, finding an auxiliary function $\widetilde{p}(\mathbf{X},\mathbf{Z})$ is an essential yet difficult part in EVI implementation. Generally speaking, this auxiliary function should satisfy the relation presented in~\eqref{Eq: Auxiliary Function} or it should satisfy
\begin{equation}
\eqs
\label{Eq: Strong Condition}
{p}(\mathbf{X},\mathbf{Z})\geq \widetilde{p}(\mathbf{X},\mathbf{Z}).
\end{equation}
It is obvious that an auxiliary function satisfies~\eqref{Eq: Strong Condition} should also satisfy~\eqref{Eq: Auxiliary Function}. Hence, the condition in~\eqref{Eq: Auxiliary Function} is named as the~\emph{weak condition} and the one in~\eqref{Eq: Strong Condition} is referred to as the~\emph{strong condition}. When using an auxiliary function to lower-bound the original objective function, the EVI will introduce a systematic gap. Generally speaking, the gap incurred by the applying weak condition is relatively smaller than that introduced by using the strong condition. Fig.~\ref{Fig: WeakAndStrongConditions} illustrates the different gaps introduced by the weak and strong conditions, respectively.

It is worthwhile to note that the auxiliary function $\widetilde{p}(\mathbf{X},\mathbf{Z})$ is not necessary to be a normalized probability density function (PDF)\footnotemark\footnotetext{Actually, an auxiliary function that satisfies the strong condition cannot be a normalized PDF, as ${p}(\mathbf{X},\mathbf{Z})$ itself is a normalized PDF.}. This will not affect the final solution since either VI or EVI will re-normalize the obtained optimal posterior distribution in the end.

In practice, in addition to the above mentioned weak or strong condition, an auxiliary function should also has a specific mathematical form so that the optimal solution in~\eqref{Eq: Optimal Approximating Solution} has the same logarithmic form as the prior distribution and the conjugate match between the prior and the posterior distributions is satisfied. This is another required condition for choosing the auxiliary function. Table~\ref{Tab: Required Conditions for EVI} lists the required conditions when implementing EVI.

Generally speaking, it is usually not feasible to find an auxiliary function that satisfies the strong condition, except that the original function $p(\mathbf{X},\mathbf{Z})$ is globally concave in terms of $\mathbf{Z}$\ \footnotemark\footnotetext{According to our experience, globally concavity holds only for Gaussian distribution. For (most of) the non-Gaussian statistical models, the original function is not globally concave.}. Compared to the strong condition, it is easy to find an auxiliary function to fulfill the weak condition, although the ordinal function $p(\mathbf{X},\mathbf{Z})$ might be partially concave with respect to part of $\mathbf{Z}$~\cite{Ma2015}. For example, the multivariate log-inverse-beta (MLIB) function in the Dirichlet distribution is~\emph{not} globally concave in terms of all of its variables. It is only relatively concave~\emph{w.r.t.} one of its variable when fixing the rest. Iteratively taking this property, an auxiliary function that satisfies the weak condition and the requirement of the mathematical form can be found so that an analytically tractable solution was derived. Moreover, the weak conditions yields smaller systematic gap. Therefore, the weak condition is more preferable in practice.

In summary, in order to apply the EVI to derive an analytically tractable solution for the Bayseisn estimation of non-Gaussian statistical models, an auxiliary function should 1) satisfies either the weak or the strong condition and 2) have the same mathematical form as the prior distribution (up to a constant difference).


\vspace{0mm}
\subsection{SLB Approximation and MLB Approximation}
\label{Sec: SLB vs MLB}
\vspace{0mm}
If we can find an auxiliary function $\widetilde{p}(\mathbf{X},\mathbf{Z})$ that contains all the variables $\mathbf{Z}$ and satisfies the aforementioned required conditions, the convergence of EVI is naturally guaranteed as this new objective function is convex or relatively convex in terms of $q_i(\mathbf{Z}_i)$~\cite{Bishop2006}. Since only one lower-bound approximation is applied to the original objective function, this approach is referred to as the single lower-bound (SLB) approximation and has been applied in,~\emph{e.g.},~\cite{Taghia2014,Ma2015}.

When dividing $\mathbf{Z}$ into $M$ disjoint groups as $\mathbf{Z} = \left\{\mathbf{Z}_1,\ldots,\mathbf{Z}_i,\ldots,\mathbf{Z}_M\right\}$, there might exist several auxiliary functions. For example, we could have $M$ auxiliary functions as
\begin{equation}
\eqs
\label{Eq: MLB}
\begin{split}
{p}(\mathbf{X},\mathbf{Z})\geq& \widetilde{p}_1(\mathbf{X},\mathbf{Z}_1)\\
\vdots\\
{p}(\mathbf{X},\mathbf{Z})\geq& \widetilde{p}_i(\mathbf{X},\mathbf{Z}_i)\\
\vdots\\
{p}(\mathbf{X},\mathbf{Z})\geq& \widetilde{p}_M(\mathbf{X},\mathbf{Z}_M).
\end{split}
\end{equation}
This approach is referred to as the multiple lower-bound (MLB) approximation. As each of the above mentioned auxiliary functions satisfies the required conditions in Sec.~\ref{Sec: Required Conditions}, the optimal solution in~\eqref{Eq: Optimal Approximating Solution} is
\begin{equation}
\eqs
\ln \widetilde{q}_i^*(\mathbf{Z}_i) = \mathbf{E}_{\mathbf{Z}\backslash_{\mathbf{Z}_i}}\left[\ln \widetilde{p}_i(\mathbf{X},\mathbf{Z}_i)\right] + \text{const}.
\end{equation}
In this case, the new objective function that maximized during each iteration is~\emph{not unique}. Hence,~\emph{there is no globally objective function that is maximized during each iteration}. The convergence cannot be theoretically guaranteed. Such procedure has been applied in~\cite{Ma2011a} and~\cite{Fan2012}. Although it is not guaranteed theoretically, the convergence was observed empirically.

Let's study a simple case with two disjoint groups in the MLB approximation. Assuming that $\mathbf{Z} = \{\mathbf{Z}_1,\mathbf{Z}_2\}$ and we have two auxiliary functions $\widetilde{p}_1(\mathbf{X},\mathbf{Z}_1)$ and $\widetilde{p}_2(\mathbf{X},\mathbf{Z}_2)$ for $\mathbf{Z}_1$ and $\mathbf{Z}_2$, respectively.
As mentioned above, two different lower-bounds are obtained as
\begin{equation}
\eqs
\label{Eq: Multiple Lower bounds}
\begin{split}
\mathcal{\widetilde{L}}_1=& \mathbf{E}_{\mathbf{Z}}\left[{\ln \widetilde{p}_1(\mathbf{X},\mathbf{Z}_1)} - \ln  q(\mathbf{Z})\right]\\
\mathcal{\widetilde{L}}_2=& \mathbf{E}_{\mathbf{Z}}\left[{\ln \widetilde{p}_2(\mathbf{X},\mathbf{Z}_2)} - \ln  q(\mathbf{Z})\right].
\end{split}
\end{equation}
If we maximize each lower-bound separately, the optimal solutions to these two disjoint groups are
\begin{subequations}
\eqs
\label{Eq: Optimal Solutions with Seperate Lower bounds}
\begin{align}
\ln \widetilde{q}_1^*(\mathbf{Z}_1) =& \mathbf{E}_{\mathbf{Z}\backslash_{\mathbf{Z}_1}}\left[\ln \widetilde{p}_1(\mathbf{X},\mathbf{Z})\right] + \text{const}\label{Eq: Optimal Solution 1 with Seperate Lower bounds}\\
\ln \widetilde{q}_2^*(\mathbf{Z}_2) =& \mathbf{E}_{\mathbf{Z}\backslash_{\mathbf{Z}_2}}\left[\ln \widetilde{p}_2(\mathbf{X},\mathbf{Z})\right] + \text{const}\label{Eq: Optimal Solution 2 with Seperate Lower bounds}.
\end{align}
\end{subequations}
With these solutions, it looks like what we are maximizing is just two times of the original lower-bound as
\begin{subequations}
\eqs
\begin{align}
2\times\mathcal{L}\geq& \mathcal{\widetilde{L}}_1+\mathcal{\widetilde{L}}_2\label{Eq: Overall Lower Bound} \\
=& \mathbf{E}_{\mathbf{Z}}\left[{\ln \widetilde{p}_1(\mathbf{X},\mathbf{Z}_1)}\right] - \mathbf{E}_{\mathbf{Z}}\left[\ln  q(\mathbf{Z})\right]\label{Eq: Lower bound 1}\\
+&\mathbf{E}_{\mathbf{Z}}\left[{\ln \widetilde{p}_2(\mathbf{X},\mathbf{Z}_2)}\right] - \mathbf{E}_{\mathbf{Z}}\left[\ln  q(\mathbf{Z})\right]\label{Eq: Lower bound 2}.
\end{align}
\end{subequations}
When performing the update strategy~\eqref{Eq: Optimal Solution 1 with Seperate Lower bounds}, we get~\eqref{Eq: Lower bound 1} to be maximized. This maximization makes the distribution of $\mathbf{Z}_1$ to be less uncertain. As $- \mathbf{E}_{\mathbf{Z}}\left[\ln  q(\mathbf{Z})\right]$ in~\eqref{Eq: Lower bound 2} is the differential entropy of $\mathbf{Z}$,~\eqref{Eq: Lower bound 2} is decreasing while~\eqref{Eq: Lower bound 1} is maximizing. It is hard to evaluate if~\eqref{Eq: Lower bound 1} changes more than~\eqref{Eq: Lower bound 2} or not. Thus, the overall lower-bound,~\emph{i.e.}, $\mathcal{\widehat{L}}_1+\mathcal{\widehat{L}}_2$ in~\eqref{Eq: Overall Lower Bound}, might decrease during some iterations. On the one hand, as the lower-bound (\emph{i.e.},~$\mathcal{\widetilde{L}}_1+\mathcal{\widetilde{L}}_2$) to the original objective function can not be guaranteed to be maximized all the time, this strategy may not promise convergence. On the other hand, if the change to~\eqref{Eq: Lower bound 1} is larger than that to~\eqref{Eq: Lower bound 2}, the convergency is still guaranteed. There is no general judgement for the convergence. It should be studied case by case. Similar arguments can be applied to the case with more than two auxiliary functions. Thus, the convergency of MLB approximation is underdetermined.
\begin{figure}[!t]
\vspace{0mm}

     \centering
     \includegraphics[width=.45\textwidth]{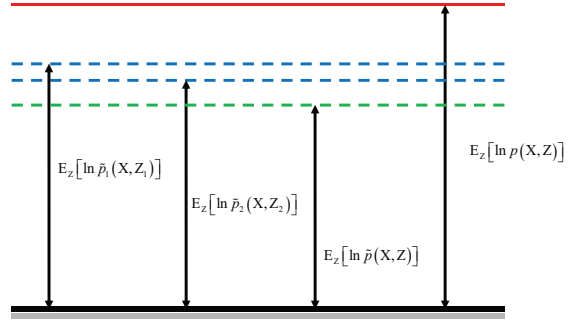}
          \caption{ \label{Fig: SLBvsMLB}\footnotesize Qualitative comparisons of SLB and MLB. For MLB, two different lower-bounds are introduced for $\mathbf{Z}_1$ and $\mathbf{Z}_2$, respectively (the blue dash lines). For SLB, there is only one lower-bound (the green dash line). The original objective function is marked with red solid line. It can be observed that the new objective function that needs to be maximized is not unique for the MLB case. Hence, the convergence is not guaranteed. A new single objective function is employed and maximized for the SLB case. Therefore, the convergence is theoretically guaranteed.}
\vspace{0mm}
\end{figure}

In summary, SLB approximation can theoretically guarantee the convergence while MLB approximation, in general, cannot promise convergence.\footnotemark\footnotetext{In practice (\emph{e.g.},~\cite{Ma2011a,Fan2012}), the EVI-based algorithm may also converge with MLB approximation. However, it is empirical result without proof.}
\vspace{0mm}
\section{Experimental Results and Discussions}
\vspace{0mm}

Recently, several EVI-based parameter estimation algorithms for non-Gaussian statistical models have been proposed. Among others, the EVI-based Bayesian BMM~\cite{Ma2011a} and the EVI-based Bayesian DMM~\cite{Fan2012} took the strong condition and the related analytically tractable solutions were derived with MLB approximation. With SLB approximation, the improved work about EVI-based Dirichlet mixture model was proposed. Regarding the non-negative matrix factorization, Hoffman~\emph{et al.} and Ma~\emph{et al.} proposed the EVI-based strategies for musical signal~\cite{Hoffman2010} and bounded support data~\cite{Ma2015}, respectively. The EVI-based von Mises-Fisher mixture model was proposed in~\cite{Taghia2014} where a structural factorization was considered. For all the aforementioned SLB approximation-based method, the weak condition is fulfilled.

In this section, we first compare the weak and strong conditions quantitatively. Secondly, we intensively compare the performance of the MLB approximation-based methods with the SLB approximation-based methods.


\subsection{Comparisons of Weak and Strong Condition }
Since Dirichlet distribution is a multivariate case of beta distribution, the EVI-based Bayesian BMM that constructs a auxiliary function with weak condition can be obtained based on the work in~\cite{Ma2014} by simply setting the dimension $K=2$. The EVI-based Bayesian BMM proposed in~\cite{Ma2011a} utilized the strong condition to choose the auxiliary function.  We compare these two different methods to demonstrate the differences between the strong and weak conditions.

Following the same notation in~\cite{Ma2011a}, we denote a multivariate Bayesian BMM with observation data $\mathbf{X}$ as
\begin{equation}
\label{BMMVector}
\eqs
\begin{split}
 f(\mathbf{X};\mathbf{\Pi},\mathbf{U},\mathbf{V}) =& \prod_{n=1}^N \sum_{i=1}^{I}\pi_i \mathrm{Beta}({x}_n;{u}_i,{v}_i),
\end{split}
\end{equation}
where $\pi_i$ is the mixture weigh for the $i$th mixture component and $\mathrm{Beta}(x;u,v)$ is the beta distribution, which can be denoted as
 \begin{equation}
\label{BetaDistr}
\eqs
\mathrm{Beta}(x;u,v) = \frac
{\Gamma(u+v)}{\Gamma(u)\Gamma(v)} x^{u-1}
(1-x)^{v-1},\ u,v >0.
\end{equation}
We consider the observation ${x}_n$ and the unobserved indication vector $\mathbf{z}_{n}$ as the \emph{complete} data. The conditional
distribution of $\mathbf{X}= \{{x}_1, \ldots, {x}_N\}$ and $\mathbf{Z}= \{\mathbf{z}_1, \ldots, \mathbf{z}_N\}$
given the latent variables $\{\mathbf{U},\mathbf{V},\mathbf{\Pi}\}$ is
\begin{equation}
\label{Conditional PDF}
\eqs
\begin{split}
f(\mathbf{X},\mathbf{Z}|\mathbf{U},\mathbf{V},\mathbf{\Pi})=&f(\mathbf{X}|\mathbf{U},\mathbf{V},\mathbf{\Pi},\mathbf{Z})f(\mathbf{Z}|\mathbf{\Pi})\\
=&f(\mathbf{X}|\mathbf{U},\mathbf{V},\mathbf{Z})f(\mathbf{Z}|\mathbf{\Pi})\\
=&\prod_{n=1}^N\prod_{i=1}^I\left[\pi_i\mathrm{Beta}({x}_n|{u}_i,{v}_i)\right]^{z_{ni}}.
\end{split}
\end{equation}
The ultimate goal is to estimate the posterior distributions of $u_{i}$, $v_{i}$, and $z_{ni}$, respectively.

In order to derive an analytically tractable solution for the posterior distributions, the most challengeable part with the EVI framework is to calculate the expectation of the bivariate log-inverse-beta (LIB) function
\begin{equation}
\eqs
\mathbf{E}_{u_{i},v_{i}}\left[\mathrm{LIB}(u_{i},v_{i})\right]= \mathbf{E}_{u_{i},v_{i}}\left[\frac{\Gamma(u_{i}+v_{i})}{\Gamma(u_{i})\Gamma(v_{i})}\right].
\end{equation}

\subsubsection{EVI-based Bayesian BMM with Weak Condition~\cite{Ma2014}}

In the Bayesian BMM with SLB approximation~\footnotemark\footnotetext{A Bayesian BMM with SLB approximation can be derived from the Bayesian DMM with SLB approximation~\cite{Ma2014} by setting the dimensions of the Dirichlet variable equal to two.}, the new objective function that we are maximizing is
\begin{equation}
\eqs
\label{Eq: EFA-SLB}
\begin{split}
&\mathbf{E}_{\mathbf{Z}}\left[\ln \widetilde{p}_{w}(\mathbf{X},\mathbf{Z})\right]\\
=&\mathcal{\widetilde{L}}_{\text{SLB}}\\
=&\ln \frac{\Gamma(\overline{u}_{i}+\overline{v}_{i})}{\Gamma(\overline{u}_{i})\Gamma(\overline{v}_{i})}\\ &+\overline{u}_{i}\left[\psi(\overline{u}_{i}+\overline{v}_{i})-\psi(\overline{u}_{i})\right] (\mathbf{E}\left[\ln u_{i}\right]-\ln \overline{u}_{i})\\
  &+ \overline{v}_{i}\left[\psi(\overline{u}_{i}+\overline{v}_{i})-\psi(\overline{v}_{i})\right] (\mathbf{E}\left[\ln v_{i}\right]-\ln \overline{v}_{i}),
\end{split}
\end{equation}
where $\overline{x}$ is the expected value of $x$ and $\psi(x)$ is the digamma function defined as $\psi(x)=\frac{\partial \ln \Gamma(x)}{\partial x}$.
This lower-bound satisfies the weak condition such that $\mathbf{E}_{\mathbf{Z}}\left[\ln p(\mathbf{X},\mathbf{Z})\right] \geq \mathbf{E}_{\mathbf{Z}}\left[\ln \widetilde{p}_{w}(\mathbf{X},\mathbf{Z})\right]$. Moreover, this lower-bound is identical for all the variables $u_{i}$, $v_{i}$, and $z_{ni}$
\subsubsection{EVI-based Bayesian BMM with Strong Condition~\cite{Ma2011a}}
For the case with strong condition, an auxiliary function $\widetilde{p}_{s}(\mathbf{X},\mathbf{Z})$ is required. In~\cite{Ma2011a}, three different auxiliary functions were derived for the variables $u_{i}$, $v_{i}$, and $z_{ni}$, respectively. To specify, for $u_{i}$, the auxiliary function is
\begin{equation}
\eqs
\label{Eq: StrongAuxiliary1}
\begin{split}
\widetilde{p}_{s_{u_i}}(\mathbf{X},\mathbf{Z})=&\ln \frac{\Gamma(\overline{u}_i+\overline{v}_i)}{\Gamma(\overline{u}_i)\Gamma(\overline{v}_i)}\\
 &+ \overline{u}_i\left[\psi(\overline{u}_i+\overline{v}_i)-\psi(\overline{u}_i)\right](\ln u_i-\ln \overline{u}_i)\\
 &+ \overline{v}_i\left[\psi(\overline{u}_i+\overline{v}_i)-\psi(\overline{v}_i)\right](\ln v_i-\ln \overline{v}_i)\\
 &+ \overline{u}_i\overline{v}_i\psi^{'}(\overline{u}_i+\overline{v}_i)(\ln u_i - \ln \overline{u}_i),
\end{split}
\end{equation}
where $\psi^{'}(x)=\frac{\partial \psi(x)}{\partial x}$. Hence, when considering $u_{i}$ as the variable, the objective function that was maximized is~\cite{Ma2011a}
\begin{equation}
\eqs
\label{Eq: EFA-MLB-1}
\begin{split}
\mathcal{\widetilde{L}}_{\text{MLB}_{u_i}}
=&\mathbf{E}_{\mathbf{Z}}\left[\widetilde{p}_{s_{u_i}}(\mathbf{X},\mathbf{Z})\right]\\
=&\ln \frac{\Gamma(\overline{u}_{i}+\overline{v}_{i})}{\Gamma(\overline{u}_{i})\Gamma(\overline{v}_{i})}\\ &+\overline{u}_{i}\left[\psi(\overline{u}_{i}+\overline{v}_{i})-\psi(\overline{u}_{i})\right] (\mathbf{E}\left[\ln u_{i}\right]-\ln \overline{u}_{i})\\
  &+ \overline{v}_{i}\left[\psi(\overline{u}_{i}+\overline{v}_{i})-\psi(\overline{v}_{i})\right] (\mathbf{E}\left[\ln v_{i}\right]-\ln \overline{v}_{i})\\
 &+ \overline{u}_{i}\cdot\overline{v}_{i}\cdot\psi^{'}(\overline{u}_{i}+\overline{v}_{i})(\mathbf{E}\left[\ln u_{i}\right] - \ln \overline{u}_{i}).
 \end{split}
\end{equation}

Similarly, due to the symmetry of $u_i$ and $v_i$, the objective function, when treating $v_i$ as the variable, is~\cite{Ma2011a}
\begin{equation}
\eqs
\label{Eq: EFA-MLB-2}
\begin{split}
\mathcal{\widetilde{L}}_{\text{MLB}_{v_i}}
=&\ln \frac{\Gamma(\overline{u}_{i}+\overline{v}_{i})}{\Gamma(\overline{u}_{i})\Gamma(\overline{v}_{i})}\\ &+\overline{u}_{i}\left[\psi(\overline{u}_{i}+\overline{v}_{i})-\psi(\overline{u}_{i})\right] (\mathbf{E}\left[\ln u_{i}\right]-\ln \overline{u}_{i})\\
  &+ \overline{v}_{i}\left[\psi(\overline{u}_{i}+\overline{v}_{i})-\psi(\overline{v}_{i})\right] (\mathbf{E}\left[\ln v_{i}\right]-\ln \overline{v}_{i})\\
 &+ \overline{u}_{i}\cdot\overline{v}_{i}\cdot\psi^{'}(\overline{u}_{i}+\overline{v}_{i})(\mathbf{E}\left[\ln v_{i}\right] - \ln \overline{v}_{i}).
 \end{split}
\end{equation}
When taking $z_{ni}$ as the only variable, the auxiliary function that proposed in~\cite{Ma2011a} is
\begin{equation}
\eqs
\label{Eq: EFA-MLB-3}
\begin{split}
&\widetilde{p}_{s_{z_{ni}}}(\mathbf{X},\mathbf{Z})\\
=&\ln \frac{\Gamma(\overline{u}_{i}+\overline{v}_{i})}{\Gamma(\overline{u}_{i})\Gamma(\overline{v}_{i})}\\ &+\overline{u}_{i}\left[\psi(\overline{u}_{i}+\overline{v}_{i})-\psi(\overline{u}_{i})\right] (\ln u_{i}-\ln \overline{u}_{i})\\
  &+ \overline{v}_{i}\left[\psi(\overline{u}_{i}+\overline{v}_{i})-\psi(\overline{v}_{i})\right] (\ln v_{i}-\ln \overline{v}_{i})\\
&+ 0.5\cdot \overline{u}_{i}^2\left[\psi^{'}(\overline{u}_{i}+\overline{v}_{i})-\psi^{'}(\overline{u}_{i})\right](\ln u_{i} - \ln \overline{u}_{i})^2\\
&+ 0.5\cdot \overline{v}_{i}^2\left[\psi^{'}(\overline{u}_{i}+\overline{v}_{i})-\psi^{'}(\overline{v}_{i})\right](\ln v_{i} - \ln \overline{v}_{i})^2\\
 &+ \overline{u}_{i}\cdot\overline{v}_{i}\cdot\psi^{'}(\overline{u}_{i}+\overline{v}_{i})(\ln u_{i} - \ln \overline{u}_{i})(\ln v_{i} - \ln \overline{v}_{i}).
 \end{split}
\end{equation}
Correspondingly, the objective function for updating the posterior distribution of $z_{ni}$ can be represented as
\begin{equation}
\eqs
\label{Eq: EFA-MLB}
\begin{split}
&\mathcal{\widetilde{L}}_{\text{MLB}_{z_{ni}}}\\
=&\mathbf{E}_{\mathbf{Z}}\left[\widetilde{p}_{s_{z_{ni}}}(\mathbf{X},\mathbf{Z})\right]\\
=&\ln \frac{\Gamma(\overline{u}_{i}+\overline{v}_{i})}{\Gamma(\overline{u}_{i})\Gamma(\overline{v}_{i})}\\ &+\overline{u}_{i}\left[\psi(\overline{u}_{i}+\overline{v}_{i})-\psi(\overline{u}_{i})\right] (\mathbf{E}\left[\ln u_{i}\right]-\ln \overline{u}_{i})\\
 & + \overline{v}_{i}\left[\psi(\overline{u}_{i}+\overline{v}_{i})-\psi(\overline{v}_{i})\right] (\mathbf{E}\left[\ln v_{i}\right]-\ln \overline{v}_{i})\\
&+ 0.5\cdot \overline{u}_{i}^2\left[\psi^{'}(\overline{u}_{i}+\overline{v}_{i})-\psi^{'}(\overline{u}_{i})\right]\mathbf{E}\left[(\ln u_{i} - \ln \overline{u}_{i})^2\right]\\
&+ 0.5\cdot \overline{v}_{i}^2\left[\psi^{'}(\overline{u}_{i}+\overline{v}_{i})-\psi^{'}(\overline{v}_{i})\right]\mathbf{E}\left[(\ln v_{i} - \ln \overline{v}_{i})^2\right]\\
 &+ \overline{u}_{i}\cdot\overline{v}_{i}\cdot\psi^{'}(\overline{u}_{i}+\overline{v}_{i})(\mathbf{E}\left[\ln u_{i}\right] - \ln \overline{u}_{i})(\mathbf{E}\left[\ln v_{i}\right] - \ln \overline{v}_{i}).
 \end{split}
\end{equation}

It has been analyzed in Sec.~\ref{Sec: Required Conditions} that both the strong condition and the weak condition incur systematic gaps. We now quantitatively compare the gaps. It is worth to note that the EVI-based Bayesian BMM with strong condition is also a MLB approximation. We focus only on the comparisons of weak and strong conditions in thie section. The comparisons about the SLB approximation with the MLB approximation will be presented in the next section.

When taking $u_i$ as the variable, the difference between the objective functions obtained via weak and strong conditions, respectively, can be calculated as
\begin{equation}
\eqs
\label{Eq: Lower bound Difference-1}
\begin{split}
\Delta\mathcal{\widetilde{L}}_{\text{SLB}\ \text{vs.}\ \text{MLB}_{u_i}}=&\mathcal{\widetilde{L}}_{\text{SLB}} - \mathcal{\widetilde{L}}_{\text{MLB}_{u_i}}\\
 =& -\bar{u}_{i}\bar{v}_{i}\psi'(\bar{u}_{i}+\bar{v}_{i})(\mathbf{E}\left[\ln v_{i}\right] - \ln \bar{v}_{i})\\
 \geq & 0,
\end{split}
\end{equation}
where we used the fact that $\psi^{'}(x)>0$ and $\ln x$ is a convex function in terms of $x$. For $v_i$, it is straightforward to show the difference is also positive by using the symmetric properties.

When comparing $\mathcal{\widetilde{L}}_{\text{SLB}} $ with $\mathcal{\widetilde{L}}_{\text{MLB}_{z_{ni}}}$, the difference is
\begin{equation}
\eqs
\label{Eq: Lower bound Difference-3}
\begin{split}
&\Delta\mathcal{\widetilde{L}}_{\text{SLB}\ \text{vs.}\ \text{MLB}_{z_{ni}}}\\
=&\mathcal{\widetilde{L}}_{\text{SLB}} - \mathcal{\widetilde{L}}_{\text{MLB}_{z_{ni}}}\\
 =&-\left\{0.5\cdot \overline{u}_{i}^2\left[\psi^{'}(\overline{u}_{i}+\overline{v}_{i})-\psi^{'}(\overline{u}_{i})\right]\mathbf{E}\left[(\ln u_{i} - \ln \overline{u}_{i})^2\right]\right.\\
&+ 0.5\cdot \overline{v}_{i}^2\left[\psi^{'}(\overline{u}_{i}+\overline{v}_{i})-\psi^{'}(\overline{v}_{i})\right]\mathbf{E}\left[(\ln v_{i} - \ln \overline{v}_{i})^2\right]\\
 &+ \left.\overline{u}_{i}\cdot\overline{v}_{i}\cdot\psi^{'}(\overline{u}_{i}+\overline{v}_{i})(\mathbf{E}\left[\ln u_{i}\right] - \ln \overline{u}_{i})(\mathbf{E}\left[\ln v_{i}\right] - \ln \overline{v}_{i})\right\}.
\end{split}
\end{equation}
It can be proved that the difference $\Delta\mathcal{\widetilde{L}}_{\text{SLB}\ \text{vs.}\ \text{MLB}_{z_{ni}}}$ is also greater than or equal to $0$. More details for this proof can be found in Appendix~\ref{Appendix-1}.

The aforementioned three positive differences indicate that the new objective function with weak condition~\cite{Ma2014} is tighter (\emph{i.e.}, closer to the original objective function) than that with strong condition~\cite{Ma2011a}. Thus, for the EVI-based Bayesian BMM, the systematic gap incurred by the weak condition is smaller than that incurred by the strong condition. This makes the weak condition more favorable in practice~\cite{Ma2014,Taghia2014,Ma2015,Hoffman2010}.

Similar analysis can be applied to the Bayesian DMM with MLB~\cite{Fan2012} and the Bayesian DMM with SLB~\cite{Ma2014}, as Dirichlet distribution is an multivariate extension of beta distribution.

\subsection{Comparisons of MLB and SLB Approximations}
In the previous section, we analyzed and compared the weak and the strong conditions for the EVI framework. Another important issue in EVI implementation is to distinguish the MLB and the SLB approximations, as the latter one can guarantee convergence but the first one may not. To this end, we compare the MLB approximation-based algorithm with the SLB approximation-based algorithm in this section.

\subsubsection{Observations of Non-convergence}
\begin{figure}[!t]
\vspace{0mm}
\psfrag{x}[][]{\tiny ${\text{Iter.}}\ \sharp$}
\psfrag{y}[][]{\tiny $\mathbf{E}_{\mathbf{Z}}\left[\ln{ p(\mathbf{X},\mathbf{Z})}-\ln{q(\mathbf{Z})}\right]$}
     \centering
          \subfigure[\sps Model A]{\includegraphics[width=.235\textwidth]{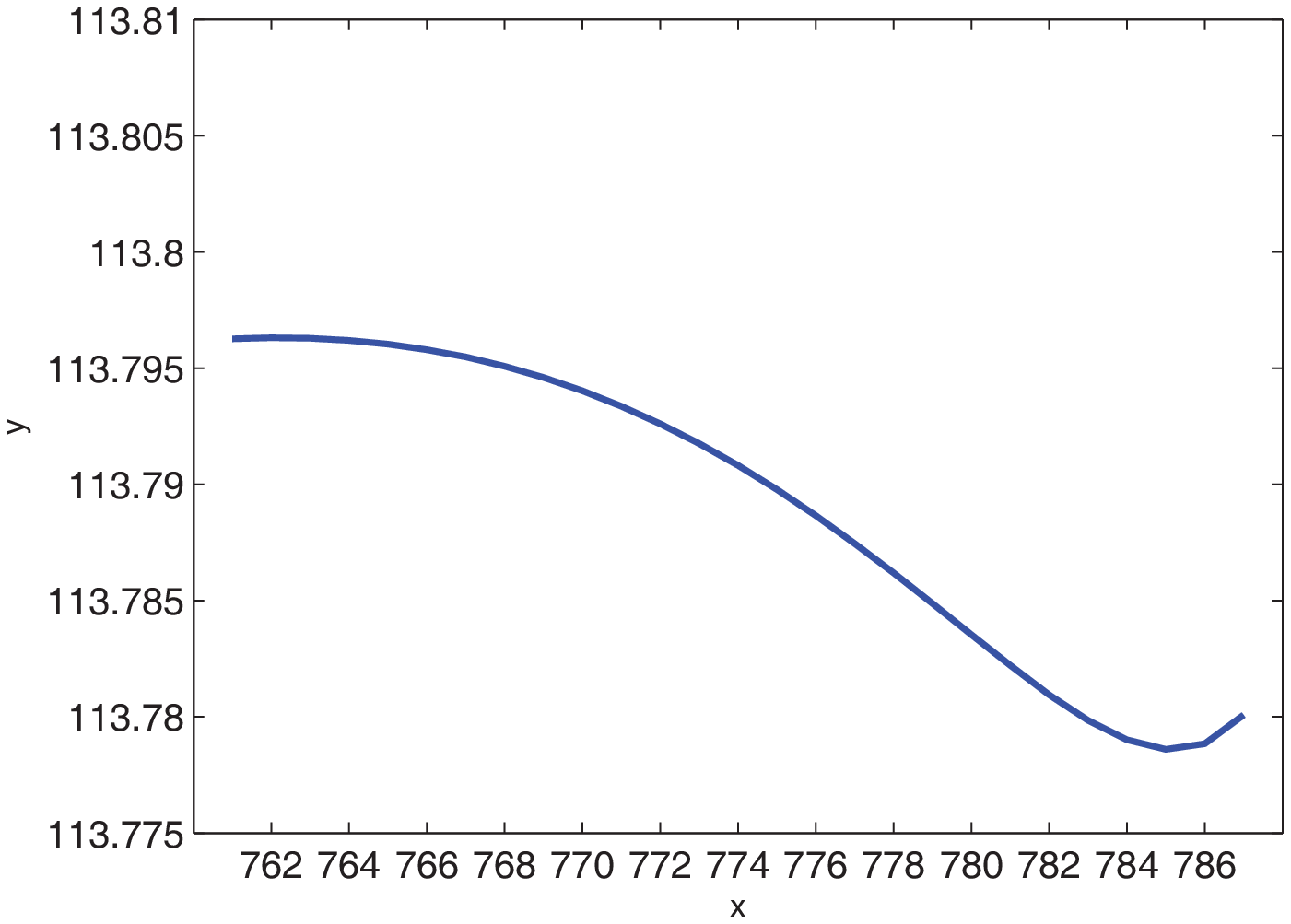}}\hspace{2mm}
          \subfigure[\sps Model B]{\includegraphics[width=.235\textwidth]{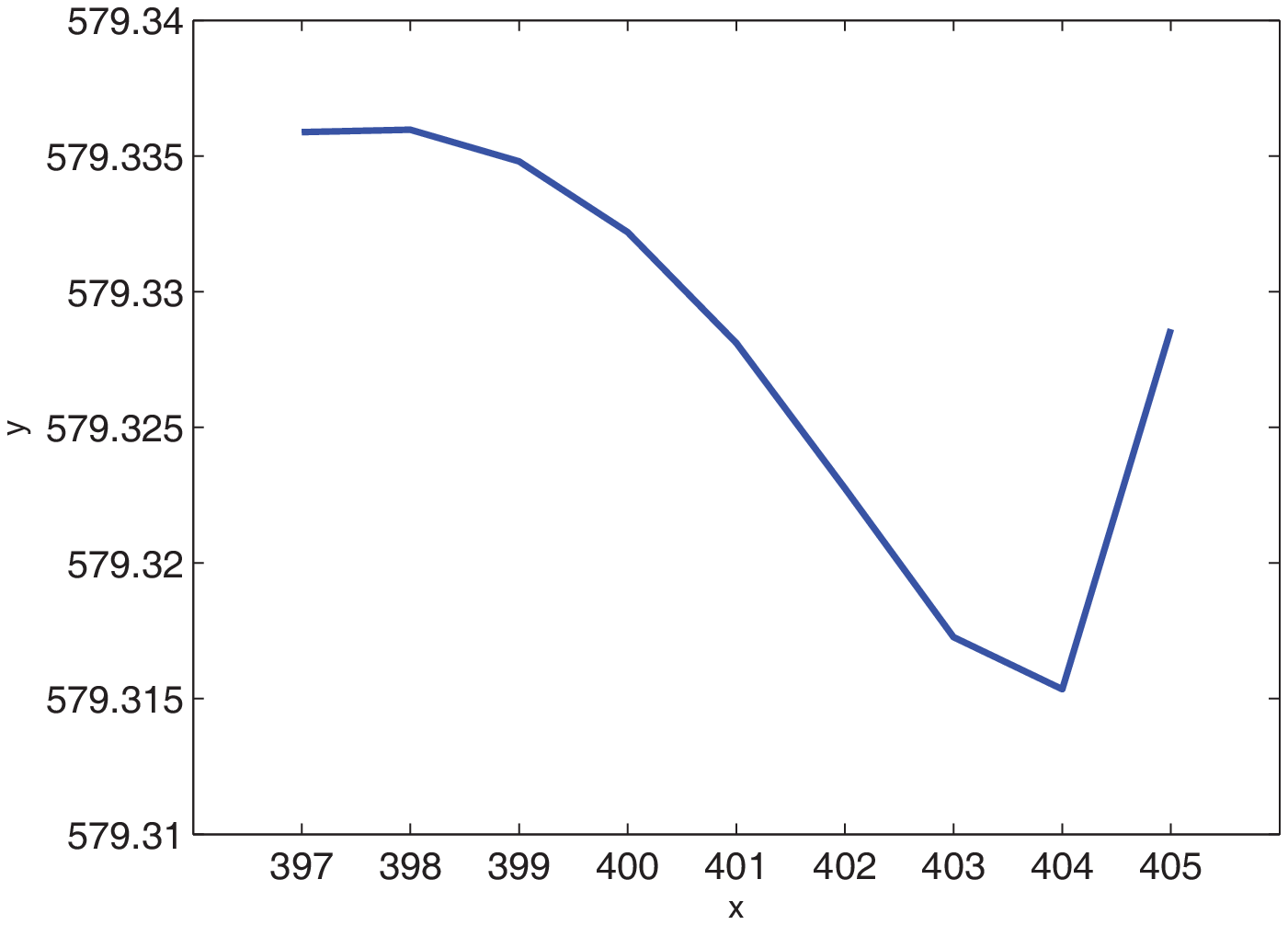}}\hspace{2mm}
\vspace{0mm}
          \caption{ \label{Fig: NonConvergence}\footnotesize Observation of decreasing values of the objective function during iterations. In principle, objective function should always increase (at least not decrease). Although this non-convergence can be observed in some of the simulation rounds ($2\sim3$ times out of $10$ rounds of simulations), this fact indicates that the MLB approximation-based method may not promise convergence. Model A is a BMM with parameter $\pi_1=0.3,\pi_2=0.7,\mathbf{u}_1=[2\ 8]^{\text{T}},\mathbf{u}_2=[15\ 4]^{\text{T}}$ and model B is a three-dimensional DMM with parameter $\pi_1=0.35,\pi_2=0.65,\mathbf{u}_1=[4\ 12\ 3]^{\text{T}}, \mathbf{u}_2=[10\ 6\ 2]^{\text{T}}$. $400$ samples were generated from each model.}
\vspace{0mm}
\end{figure}

As discussed in Sec.~\ref{Sec: SLB vs MLB}, the convergence of the MLB method is not guaranteed. We ran the MLB approximation-based Bayesian BMM algorithm~\cite{Ma2011a} and Bayesian DMM algorithm~\cite{Fan2012}, respectively, and monitored the value of the objective function during each iteration. It can be observed that, for some rounds of simulations~\footnotemark\footnotetext{Here, one simulation round means that we ran the estimation algorithm until it stops according to some criterion.}, the objective function is~\emph{decreasing} during some iterations. This phenomenon has been observed for several times, both for BMM and DMM. Figure~\ref{Fig: NonConvergence} lists the decreasing objective function values and the corresponding iterations. For the SLB approximation-based Bayesian BMM and Bayesian DMM~\cite{Ma2014}, the monitored objective function was always increasing until converging. The observation of non-convergence demonstrates that the convergence with MLB approximation is underdetermined.
%
\subsubsection{Comparisons of Estimation Accuracy}

In this section, we compare the MLB approximation with the SLB approximation quantitatively. With a known BMM or DMM, $2,000$ samples were generated, respectively. The above-mentioned Bayesian estimation algorithms were applied to estimate the posterior distributions, respectively. We calculated the original variational objective function in~\eqref{Eq: Lowerbound} to examine which approximation is better. With the obtained posterior distribution $q^*(\mathbf{Z})$, the original variational objective function is calculated numerically by sampling method. Hence,we got two different values, ${\mathcal{{L}}_{\text{SLB}}}$ and $\mathcal{{L}}_{\text{MLB}}$, from the SLB approximation and the MLB approximation, respectively. Larger value means closer lower-bound approximation. In addition to this, we also measure the estimation accuracy by the KL divergence of the estimated PDF from the true one as
$\text{KL}(p(\mathbf{X}|\boldsymbol\Theta)\|p(\mathbf{X}|\boldsymbol{\widehat{\Theta}}))$,
where $\boldsymbol\Theta$ is the true parameter vector and $\boldsymbol{\widehat{\Theta}}$ is the estimated one. Similarly, we numerically calculated $\text{KL}_{\text{SLB}}$ and $\text{KL}_{\text{MLB}}$ from the SLB and MLB approximations~\footnotemark\footnotetext{For the MLB approximation, we only take those simulation rounds that always converge into consideration.}, respectively. The smaller the KL divergence is, the more accurate the estimation is.

\begin{figure}[!t]
\vspace{0mm}
\psfrag{x}[][]{\tiny ${\text{SLB}}$}
\psfrag{y}[][]{\tiny ${\text{MLB}}$}
     \centering
          \subfigure[\label{Subfig: AN10}\sps Model A]{\includegraphics[width=.235\textwidth]{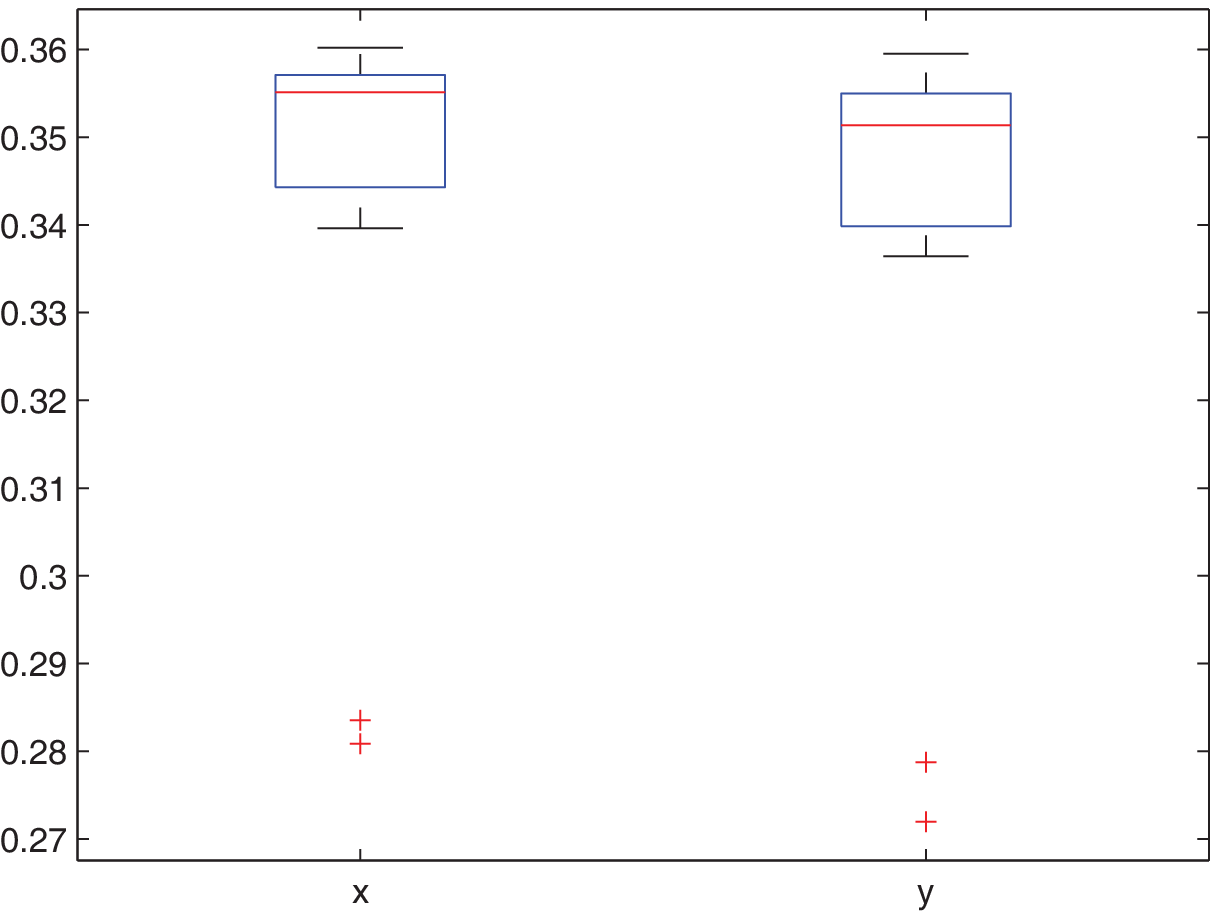}}\hspace{2mm}
          \subfigure[\sps Model B]{\includegraphics[width=.235\textwidth]{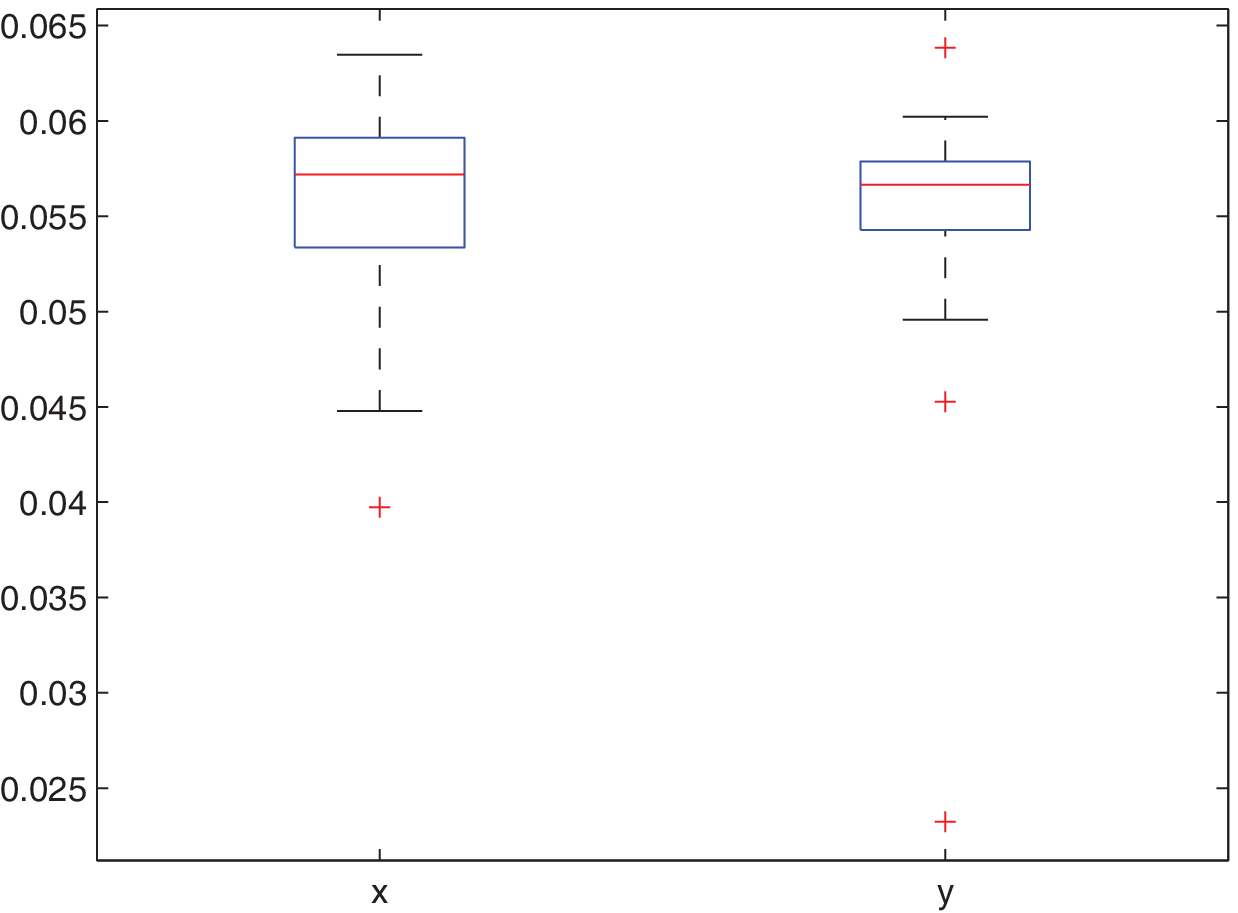}}\hspace{2mm}
\vspace{0mm}
          \caption{ \label{Fig: BMM}\footnotesize Comparisons of the original objective functions in Bayesian BMM. The central mark is the median, the edges are the $25^{th}$ and $75^{th}$ percentiles. The outliers are marked individually. Model settings are the same as Table~\ref{Tab: BMM}.}
\vspace{0mm}
\end{figure}

For Bayesian BMM, the comparisons are presented in Table~\ref{Tab: BMM} and Figure~\ref{Fig: BMM}. The simulations were run $20$ rounds and the mean values are reported. The comparisons of the Bayesian DMM via SLB~\cite{Ma2014} and MLB~\cite{Fan2012} approximations are illustrated in Table~\ref{Tab: DMM} and Figure~\ref{Fig: DMM}. It can be observed that, for both Bayesian BMM and Bayesian DMM, the SLB approximation yields higher objective function value than the MLB approximation. Meanwhile, the KL divergences obtained by the SLB approximation are all smaller than those obtained by the MLB. These facts demonstrate that the SLB approximation is superior to the MLB approximation.

\begin{table}[!t]
  \vspace{0mm}
  \caption{\label{Tab: BMM} \footnotesize Comparisons of the objective functions for Bayesian BMM. }
 \centering
\sps
 \begin{tabular}{|@{}c@{}|c|c|c|c|}
  \hline
 \ Model\ \ &Parameters & \ $\mathcal{L}_{\text{SLB}}- \mathcal{L}_{\text{MLB}}$\ &\ $\text{KL}_{\text{SLB}}-\text{KL}_{\text{MLB}}$\ \\[.1mm]
 \hline
 \hline
\multirow{2}{*}{A}&$\pi_1 = 0.3, u_1 = 2,v_1 = 8$  & \multirow{2}{*}{$3.6\times 10^{-3}$}& \multirow{2}{*}{$-2.8\times 10^{-3}$} \\[.1mm]
&$\pi_2 = 0.7, u_2 = 15,v_2 = 4$  & & \\[.1mm]
 \hline
\multirow{3}{*}{B}&$\pi_1 = 0.3, u_1 = 10,v_1 = 2$  & \multirow{3}{*}{$1.3\times 10^{-3}$} & \multirow{3}{*}{$-0.58\times 10^{-3}$}\\[.1mm]
&$\pi_2 = 0.4, u_2 = 2,v_2 = 12$  &  &\\[.1mm]
&$\pi_3 = 0.3, u_3 = 10,v_3 = 10$  &  &\\[.1mm]
\hline\end{tabular}\\
\end{table}

\vspace{0mm}
\section{Conclusions}
\vspace{0mm}
The extended variational inference (EVI) framework can be applied in efficiently estimation of non-Gaussian statistical models. We discussed and summarized the required conditions for selection of the auxiliary functions in the EVI framework. Moreover, we also analyzed and compared the single lower-bound (SLB) approximation and the multiple lower-bounds (MLB) approximation. Theoretical analysis showed that the weak condition, in general, can incur smaller systematic gap than the strong condition. Hence, the weak condition is more preferable in practice. Furthermore, quantitative evaluations based on Bayesian beta mixture model and Bayesian Dirichlet mixture model demonstrated that the SLB approximation can theoretically guarantee convergence and is superior to the MLB approximation.


\end{document}